\documentclass[sigconf]{aamas} 

\usepackage{balance} 

\setcopyright{ifaamas}
\acmConference[AAMAS '24]{Proc.\@ of the 23rd International Conference
on Autonomous Agents and Multiagent Systems (AAMAS 2024)}{May 6 -- 10, 2024}
{Auckland, New Zealand}{N.~Alechina, V.~Dignum, M.~Dastani, J.S.~Sichman (eds.)}
\copyrightyear{2024}
\acmYear{2024}
\acmDOI{}
\acmPrice{}
\acmISBN{}

\acmSubmissionID{427}

\title[AAMAS-2024 Formatting Instructions]{High-Level, Collaborative Task Planning Grammar and Execution for Heterogeneous Agents}

\author{Amy Fang}
\affiliation{
  \institution{Cornell University}
  \city{Ithaca}
  \state{NY}
  \country{USA}}
\email{axf4@cornell.edu}

\author{Hadas Kress-Gazit}
\affiliation{
  \institution{Cornell University}
  \city{Ithaca}
  \state{NY}
  \country{USA}}
\email{hadaskg@cornell.edu}

\begin{abstract}
We propose a new multi-agent task grammar to encode collaborative tasks for a team of heterogeneous agents that can have overlapping capabilities. The grammar allows users to specify the relationship between agents and parts of the task without providing explicit assignments or constraints on the number of agents required. We develop a method to automatically find a team of agents and synthesize correct-by-construction control with synchronization policies to satisfy the task. 
We demonstrate the scalability of our approach through simulation and compare our method to existing task grammars that encode multi-agent tasks.
\end{abstract}


\newcommand{\BibTeX}{\rm B\kern-.05em{\sc i\kern-.025em b}\kern-.08em\TeX}

\usepackage{amsmath}
\usepackage[geometry]{ifsym}
\usepackage{graphicx}
\newsavebox\mybox

\usepackage{ifsym}

\usepackage{caption}
\usepackage{subcaption}
\usepackage{array}
\usepackage{enumitem}
\usepackage{dblfloatfix}

\usepackage{enumerate}
\usepackage{algorithmic}
\usepackage[ruled,linesnumbered,noend]{algorithm2e}

\SetArgSty{textnormal}

\newcommand{\APbinding}{AP_{\psi}}
\newcommand{\APltl}{AP_{\varphi}}
\newcommand{\ltlpsi}{LTL$^\psi$}
\newcommand{\buchi}{B{\"u}chi }

\newcommand{\syntline}{\: | \:}
\newcommand{\X}{\bigcirc}
\newcommand{\G}{\Box}
\newcommand{\F}{\Diamond}
 
\usepackage{amsthm}
\theoremstyle{plain}

\newtheoremstyle{slanted}
  {0.25em plus .2em}
  {0.25em plus .2em}
  {}
  {}
  {\bfseries}
  {.}
  {0.5em}
  {}

\theoremstyle{slanted}

\theoremstyle{remark}

\newtheoremstyle{exam}
  {0.25em plus .2em}
  {0.25em plus .2em}
  {}
  {}
  {\slshape}
  {.}
  {0.5em}
  {}

\theoremstyle{exam}

\usepackage{color}
\usepackage{xcolor}

\begin{document}


\pagestyle{fancy}
\fancyhead{}


\maketitle 


\section{Introduction}

Agents working together to achieve common goals have a variety of applications, such as warehouse automation or disaster response. Multi-agent tasks have been defined in different ways in the scheduling and planning literature. For example, in multi-agent task allocation ~\cite{Jia2013, Gerkey2004, Korsah2013} and coalition formation \cite{Xu2015, Li2009}, each task is a single goal with an associated utility. Individual agents or agent teams then automatically assign themselves to a task based on some optimization metric. Swarm approaches \cite{Schmickl2006, Wang2020} consider emergent behavior of an agent collective as the task, for example, aggregation or shape formation. 

Recently, formal methods, such as temporal logics for task specifications and correct-by-construction synthesis, have been used to solve different types of multi-agent planning tasks \cite{Ulusoy2012, Schillinger2018, Chen2021}. Tasks written in temporal logic, such as Linear Temporal Logic (LTL), allow users to capture complex tasks with temporal constraints. Existing work has extended LTL \cite{Luo2022,Sahin2017} and Signal Temporal Logic \cite{Leahy2022} to encode tasks that require multiple agents.


In this paper, we consider tasks that a team of heterogeneous agents are required to collaboratively satisfy. For instance, consider a precision agriculture scenario in which a farm contains agents with different on-board sensors to monitor crop health. The user may want to take a moisture measurement in one region, and then take a soil sample of a different region. Depending on the agents' sensors and sensing range, the agents may decide to collaborate to satisfy the task. For example, one agent may perform the entire task on its own if it has both a moisture sensor and an arm mechanism to pick up a soil sample and can move between the two regions. However, another possible solution is for two agents to team up so that one takes a moisture measurement and the other picks up the soil. Existing task grammars \cite{Luo2022, Sahin2017, Leahy2022} capture tasks such as the above by providing explicit constraints on the types or number of agents for each part of the task, i.e. the task must explicitly encode whether it should be one agent, two agents, or either of these options. In this paper, we create a task grammar and associated control synthesis that removes the need to a priori decide on the number of agents necessary to accomplish a task, allowing users to focus solely on the actions required to achieve the task (e.g. ``take a moisture measurement and then pick up a soil sample, irrespective of which or how many agents perform which actions"). 

Our task grammar has several unique aspects. 
{First, this grammar enables the interleaving of agent actions, alleviating the need for explicit task decomposition in order to assign agents to parts of the task.} Second, rather than providing explicit constraints on the types or number of agents for each part of the task, the task encodes, using the concept of \textit{bindings} (inspired by~\cite{Luo2022}), the overall relationship between agent assignments and team behavior; we can require certain parts of the task to be satisfied by the same agent without assigning the exact agent or type of agent \textit{a priori}. {Lastly, the grammar allows users to make the distinction between the requirements ``for all agents'' and ``at least one agent''}. 
Given these types of tasks, agents autonomously determine, based on their capabilities, which parts of the task they can and should do for the team to satisfy the task.

Tasks may require collaboration between different agents. Similar to \cite{Kloetzer2010,Tumova2016,Chen2011}, to ensure the actions are performed in the correct order, our framework takes the corresponding synchronization constraints into account while synthesizing agent behavior; agents must wait to execute the actions together. In our approach, execution of the synchronous behavior for each agent is decentralized; agents carry out their plan and communicate with one another when synchronization is necessary.

Depending on the task and the available agents, there might be different teams (i.e., subsets of the agent set) that can carry out the task; our algorithm for assigning a team and synthesizing behavior for the agents finds the largest team of agents that satisfies the task. This means that the team may have redundancies, i.e. agents can be removed while still ensuring the overall task is satisfied. This is beneficial both for robustness and optimality; the user can choose a subset of the team (provided that all the required bindings are still assigned) to optimize different metrics, such as cost or overall number of agents. 


\textbf{Related work:} 
One way to encode tasks is to first decompose them into independent sub-tasks and then allocate them to the agents. 
For example, \cite{Schillinger2018, Faruq2018} address finite-horizon tasks for multi-agent teams. The authors first automatically decompose a global automaton representing the task into independent sub-tasks. To synthesize control policies, the authors build product automata for each heterogeneous agent. Each automaton is then sequentially linked using switch transitions to reduce state-space explosion in synthesizing parallel plans. 
In our prior work \cite{Fang2022}, we address infinite-horizon tasks that have already been decomposed into sub-tasks. Given a new task, we proposed a decentralized framework for agents to automatically update their behavior based on a new task and their existing tasks, allowing agents to interleave the tasks. 



The works discussed above make the critical assumption that tasks are independent, i.e. agents do not collaborate with one another. One approach to including collaborative actions is to explicitly encode the agent assignments in the tasks.
To synthesize agent control for these types of tasks, in \cite{Tumova2016}, the authors construct a reduced product automaton in which the agents only synchronize when cooperative actions are required. The work in \cite{Kantaros2020} proposes a sampling-based method that approximates the product automaton of the team by building trees incrementally while maintaining probabilistic completeness. In this paper, we consider the more general setting in which agents may need to collaborate with each other, but are not given explicit task assignments \textit{a priori}.

Rather than providing predetermined task assignments, another approach for defining collaborative tasks is to capture information about the number and type of agents needed for parts of the specification. For example, \cite{Sahin2017} imposes constraints on the number of agents necessary in regions using counting LTL. \cite{Leahy2022} uses Capability Temporal Logic to encode both the number and capabilities necessary in certain abstracted locations in the environment and then formulates the problem as a MILP to find an optimal teaming strategy. The authors of \cite{Luo2022} introduce the concept of induced propositions, where each atomic proposition not only encodes information about the number, type of agents, and target regions, but also has a connector that binds the truth of certain atomic propositions together. To synthesize behavior for the agents, they propose a hierarchical approach that first constructs the automaton representing the task and then decomposes the task into possible sub-tasks. The temporal order of these sub-tasks is captured using partially ordered sets and are used in the task allocation problem, which is formulated as a MILP.


Inspired by \cite{Luo2022} and the concept of induced propositions, we create a task grammar that includes information about how the atomic propositions are related to one another, which represents the overall relationship between agents and task requirements. Unlike \cite{Luo2022}, which considers navigation tasks in which the same set of agents of a certain type may need to visit different regions, we generalize these tasks to any type of abstract action an agent may be able to perform. In addition, a key assumption we relax is that we do not require each agent to be only categorized as one type. As a result, agents can have overlapping capabilities. To our knowledge, no other grammars have been proposed for these generalized types of multi-agent collaborative tasks.

\textbf{Contributions:} We propose a task description and control synthesis framework for heterogeneous agents to satisfy collaborative tasks. Specifically, we present a new, LTL-based task grammar for the formulation of collaborative tasks, and provide a framework to form a team of agents and synthesize control and synchronization policies to guarantee the team satisfies the task. We demonstrate our approach in simulated precision agriculture scenarios.



\section{Preliminaries}

\subsection{Linear Temporal Logic}\label{sec:ltl}

LTL formulas are defined over a set of atomic propositions $AP$, where $\pi \in AP$ are Boolean variables \cite{EMERSON1990}. We abstract agent actions as atomic propositions. For example, $UV$ captures an agent taking UV measurement.

\textbf{Syntax: } An LTL formula is defined as:
\[
    \varphi ::=  \pi  \ |  \ \neg\varphi  \ | \ \varphi \vee \varphi \ | \ \bigcirc \varphi \ | \ \varphi \ \mathcal{U} \  \varphi
\]

\noindent where $\neg$ (``not") and $\vee$ (``or") are Boolean operators, and $\bigcirc$ (``next") and  $\mathcal{U}$ (``until") are temporal operators.
From these operators, we can define: conjunction $\varphi \wedge \varphi$, implication $\varphi \Rightarrow \varphi$, eventually $\Diamond \varphi = \text{True} \ \mathcal{U} \ \varphi $, and always $\Box \varphi=\neg\Diamond \neg\varphi$.

\textbf{Semantics:} The semantics of an LTL formula $\varphi$ are defined over an infinite trace $\sigma = \sigma(0)\sigma(1)\sigma(2)...$, where $\sigma(i)$ is the set of true $AP$ at position $i$. We denote that $\sigma$ satisfies LTL formula  $\varphi$ as $\sigma \models \varphi$. 

Intuitively, $\Diamond \varphi$ is satisfied if there exists a $\sigma(i)$ in which $\varphi$ is true. $\Box \varphi$ is satisfied if $\varphi$ is true at every position in $\sigma$. To satisfy $\varphi_1 \ \mathcal{U} \ \varphi_2$, $\varphi_1$ must remain true until $\varphi_2$ becomes true. 
See \cite{EMERSON1990} for the full semantics. 
\subsection{\buchi Automata} \label{sec:buchi_standard}

An LTL formula $\varphi$ can be translated into a Nondeterministic \buchi Automaton that accepts infinite traces if and only if they satisfy $\varphi$. A \buchi automaton is a tuple $\mathcal{B}= (Z, z_0, \Sigma_{\mathcal{B}}, \delta_{\mathcal{B}}, F)$, where $Z$ is the set of states, $z_0 \in Z$ is the initial state, $\Sigma_{\mathcal{B}}$ is the input alphabet, $\delta_{\mathcal{B}}: Z \times \Sigma_{\mathcal{B}} \times Z$ is the transition relation, and $F \subseteq Z$ is a set of accepting states.
An infinite run of $\mathcal{B}$ over a word $w = w_1 w_2 w_3 ...$, $w_i\in \Sigma_{\mathcal{B}}$ is an infinite sequence of states $z = z_0 z_1 z_2...$ such that $(z_{i-1}, w_i, z_{i}) \in \delta_{\mathcal{B}}$. A run is accepting if and only if Inf($z$) $\cap \ F \neq \emptyset$, where Inf($z$) is the set of states that appear in $z$ infinitely often \cite{Baier2008}.

\subsection{Agent Model} 


Following \cite{Fang2022}, we create an abstract model for each agent based on its set of capabilities. A capability is a weighted transition system $\mathcal{\lambda} = (S, s_0, AP, \Delta, L, W)$, where $S$ is a finite set of states, $s_0 \in S$ is the initial state, $AP$ is the set of atomic propositions, $\Delta \subseteq S \times S$ is a transition relation where for all $s \in S$, $\exists s' \in S$ such that $(s,s') \in \Delta$,  $L: S \rightarrow 2^{AP}$ is the labeling function such that $L(s)$ is the set of propositions that are true in state $s$, and $W: \Delta \rightarrow \mathbb{R}_{\ge 0}$ is the cost function assigning a weight to each transition. Since we are considering a group of heterogeneous agents, agent $j$ has its own set of $k$ capabilities $\Lambda_j =  \{\lambda_1, ..., \lambda_k\}$.



\begin{figure}
    \centering            \includegraphics[width=0.9\columnwidth]{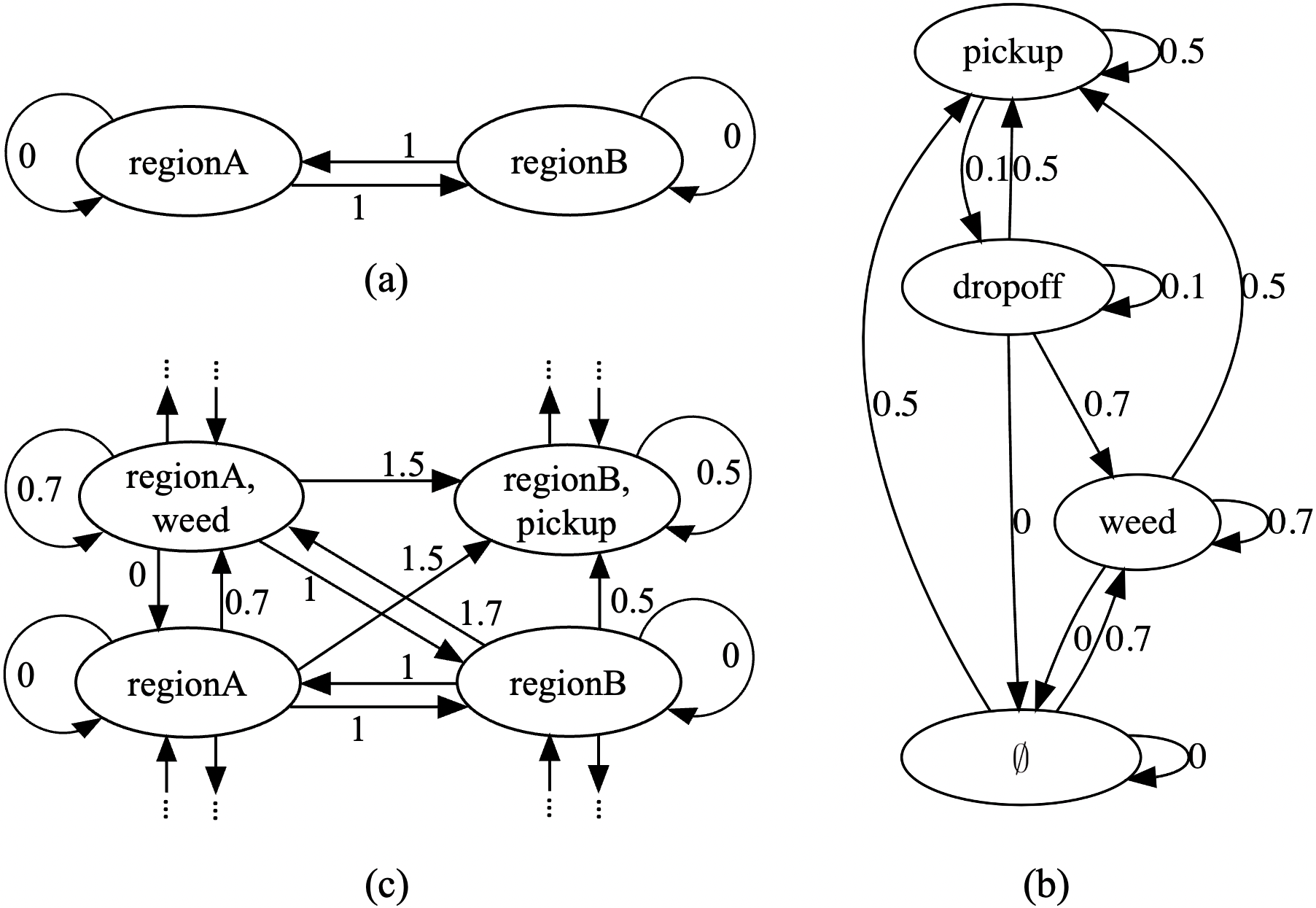}
\caption{Agent partial model: (a) $\lambda_{\mathit{area}}$ (b) $\lambda_{\mathit{arm}}$ (c) $A_{green}$
}
\label{fig:robot}
\end{figure}

An agent model $A_j$ is the product of its capabilities: $A_j = \lambda_1 \times... \times \lambda_k$ such that $A_j=(S, s_0, AP_j, \gamma, L, W)$, where $S = S_1 \times ... \times S_k$ is the set of states, $s_0 \in S$ is the initial state, $AP_j = \bigcup_{i=1}^k AP_i$
is the set of propositions, $\gamma \subseteq S \times S$ is the transition relation such that $(s,s')\in \gamma$, where $s = (s_1, ..., s_k), s' = (s'_1, ..., s'_k)$, if and only if for all $i=\{1,...,k\}, (s_i, s'_i) \in \Delta_i$, $L:S \rightarrow 2^{AP_j}$ is the labeling function where $L(s) =\bigcup_{i=1}^k L_i(s_i)$, and $W: \gamma \rightarrow \mathbb{R}_{\ge 0}$ is the cost function that combines the costs of the capabilities. 
Fig. \ref{fig:robot}c depicts a snippet of an agent model where we treat the cost as additive. Fig. \ref{fig:robot}a represents the agent's sensing area $\lambda_{area}$; the agent can orient its sensors to take measurements in different regions of a partitioned workspace (in this case, regions A and B). 
Fig. \ref{fig:robot}b represents the agent's robot manipulator, which can pick up and drop off soil samples, as well as pull weeds. 
 

\section{Task Grammar - \ltlpsi} 

{We define the task grammar  \ltlpsi \ that includes atomic propositions that abstract agent action, logical and temporal operators, as in LTL, and bindings that connect actions to specific agents; any action labeled with the same binding must be satisfied by the same agent(s) (the actual value of the binding is not important). 
We define a task recursively over LTL and binding formulas. } 
\begin{flalign}
    \psi &:= \rho \syntline \psi_1 \vee \psi_2 \syntline \psi_1 \wedge \psi_2 \\
     \varphi &:=  \pi  \ |  \ \neg\varphi  \ | \ \varphi \vee \varphi \ | \bigcirc \varphi \ | \ \varphi \ \mathcal{U} \  \varphi \\
     \varphi^{\psi} \!&:= \varphi^{\psi} \! \syntline
     \neg(\varphi^{\psi}) \syntline
    \! \varphi_1^{\psi_1} \!\! \wedge \! \varphi_2^{\psi_2} | \
    \! \varphi_1^{\psi_1} \!\! \vee \! \varphi_2^{\psi_2} \! \syntline 
    \! \! \X \! \varphi^{\psi} \! \syntline
     \varphi_1^{\psi_1} \mathcal{U} \varphi_2^{\psi_2} \! \syntline 
     \G \varphi^{\psi}\hspace{-3mm}
\end{flalign} 


\noindent where $\psi$, the \textit{binding formula}, is a Boolean formula excluding negation over $\rho\in\APbinding$, and $\varphi$ is an LTL formula. An \ltlpsi \ formula consists of conjunction, disjunction, and temporal operators; we define  eventually as $\F \varphi^{\psi} = \text{True} \ \mathcal{U} \ \varphi^{\psi}$. An example of an \ltlpsi \: formula is shown in Eq. \ref{eq:task}.

\textbf{Semantics: }   
The semantics of an \ltlpsi\ formula $\varphi^{\psi}$ are defined over $\sigma$ and $R$; $\sigma = \sigma_1\sigma_2...\sigma_n$ is the team trace where $\sigma_{j}$ is agent $j$'s trace, and $\forall i, \sigma(i) = \sigma_{1}(i)\sigma_{2}(i)...\sigma_{n}(i)$. 
$R = \{r_{1}, r_{2},..., r_{n}\}$ is the set of binding assignments, where $r_j \in R $ is the set of $\APbinding$ that are assigned to agent $j$. Once a team is established, $R$ is constant, i.e. an agent's binding assignment does not change throughout the task execution. 
For example, $r_{1} = \{2,3\}, r_{2} = \{1\}$ denotes that agent 1 is assigned bindings 2 and 3, and agent 2 is assigned binding 1. 

Given $n$ agents and a set of binding propositions $\APbinding$, we define the function $\zeta: \psi \rightarrow 2^{2^{\APbinding}}$ such that $\zeta(\psi)$ is the set of all possible combinations of $\rho$ that satisfy $\psi$. For example, $\zeta \bigl( (1 \vee 2) \wedge 3 \bigr) = \{ \{1,3\},\{2,3\}, \{1,2,3\}\}$.

The semantics of \ltlpsi are:
\begin{itemize}[leftmargin=*]
\setlength\itemsep{0.6em}
    \item $(\sigma(i), R)\! \models \varphi^{\psi}$ iff $\exists K \in \zeta(\psi)$ s.t. ($ K \subseteq \bigcup\limits_{p=1}^{n} r_{p}$) and ($\forall j$ s.t. $K \cap r_j \neq \emptyset$, $\sigma_j(i) \models \varphi$)
    
    \item $(\sigma(i), R)\! \models (\neg \varphi)^{\psi}$ iff $\exists K \in \zeta(\psi)$ s.t. ($ K \subseteq \bigcup\limits_{p=1}^{n} r_{p}$) and ($\forall j$ s.t. $K \cap r_j \neq \emptyset$, $\sigma_j(i) \not  \models \varphi$)

    \item $(\sigma(i), R)\! \models \neg(\varphi^{\psi})$ iff $\exists K \in \zeta(\psi)$ s.t. ($ K \subseteq \bigcup\limits_{p=1}^{n} r_{p}$) and ($\exists j$ s.t. $K \cap r_j \neq \emptyset$, $\sigma_j(i) \not\models \varphi$)
    
    \item $(\sigma(i),\!R)\!\!\models\!\! \varphi_1^{\psi_1} \!\wedge\varphi_2^{\psi_2}$ iff $(\sigma(i),\!R)\!\!\models \! \varphi_1^{\psi_1}$and $(\sigma(i),\!R)\! \models\!\varphi_2^{\psi_2}$
    
    \item $(\sigma(i),\!R)\! \models\! \varphi_1^{\psi_1}\!\vee\! \varphi_2^{\psi_2}$ iff $(\sigma(i),\!R)\! \models\!\varphi_1^{\psi_1}$or $(\sigma(i), R)\! \models \varphi_2^{\psi_2}$
    

    
    \item $(\sigma(i), R) \models \X \varphi^{\psi}$ iff $\sigma(i+1), R \models \varphi^{\psi}$
    

    \item $(\sigma(i), R)\! \models \varphi_1^{\psi_1} \mathcal{U} \varphi_2^{\psi_2}$ iff $\exists \ell \geq i$ s.t. $(\sigma(\ell), R) \models \varphi_2^{\psi_2}$ and $\forall i \leq k < \ell, (\sigma(k), R) \models \varphi_1^{\psi_1}$
    
      \item $(\sigma(i), R) \models \G \varphi^{\psi}$ iff $\forall \ell>i, (\sigma(\ell), R) \models \varphi^{\psi}$


\end{itemize}

Intuitively, 
the behavior of an agent team and their respective binding assignments satisfy $\varphi^{\psi}$ if there exists a possible binding assignment in $\zeta(\psi)$ in which all the bindings are assigned to (at least one) agent, and the behavior of all agents with a relevant binding assignment satisfy $\varphi$.
An agent can be assigned more than one binding, and a binding can be assigned to more than one agent.

\noindent \textbf{Remark 1.}
    For the sake of clarity in notation, $\neg \varphi^\psi$ is equivalent to $(\neg \varphi)^\psi$. For example, $\neg pickup^1 \triangleq (\neg pickup)^1$.

\noindent \textbf{Remark 2.}
    Note the subtle but important difference between $(\neg \varphi)^{\psi}$ and $\neg (\varphi^{\psi})$. Informally, the former requires all agents with binding assignments that satisfy $\psi$ to satisfy $\neg \varphi$; the latter requires the formula $\varphi^{\psi}$ to be violated, meaning that at least one agent's trace violates $\varphi$, i.e. satisfies $\neg\varphi$.

\noindent \textbf{Remark 3.}
    Unique to \ltlpsi is the ability to encode both tasks that include constraints on all agents or on at least one agent;
    ``For all agents'' is captured by 
    $\varphi^\psi$; 
    ``at least one agent'' is encoded as $\neg((\neg \varphi)^{\psi})$, which captures ``at least one agent assigned a binding in $K \in \zeta(\psi)$ satisfies $\varphi$''. This allows for multiple agents to be assigned the same binding, but only one of those agents is necessary to satisfy $\varphi$. This can be particularly useful in tasks with safety constraints; for example, we can write $\neg (\neg region_A^1) \Rightarrow (region_A \wedge visual)^2$, which says ``if any agent assigned binding 1 is in region A, all agents assigned binding 2 must take a picture of the region.'' 


\textbf{\textit{Example.}}
 Let $\APbinding=\{1,2,3\}$, $\APltl=\{region_A, region_B, pickup,$ \\$thermal, visual, moisture, UV\}$, and $\varphi^\psi = \varphi^\psi_1 \wedge \varphi^\psi_2$, where
\begin{subequations}\label{eq:task}
\begin{align}
&\varphi^\psi_1 \! = \! \Diamond((region_B \!\wedge\! moisture \!\wedge\! UV)^{2 \wedge 3} \!\!\wedge \! (region_A \!\wedge\! pickup)^{1}) \label{eq:task_part1} \\ 
&\varphi^\psi_2 =\neg pickup^{1} \ \mathcal{U} \ (region_A \nonumber\\
&\;\;\;\;\;\;\;\;\;\;\;\; \wedge ((thermal \vee visual) \wedge \neg(thermal \wedge visual)))^{2} 
\label{eq:task_part2}
\end{align}

\end{subequations}
$\varphi^\psi_1$ captures ``Agent(s) assigned bindings 2 and 3 should take a moisture measurement and a UV measurement in the region B at the same time that agent(s) assigned binding 1 picks up a soil sample in region A." $\varphi^\psi_2$ captures ``Before the soil sample can be picked up, agent(s) assigned binding 2 needs to either take a thermal image or a visual image (but not both) of region A.''

Note that, since multiple bindings can be assigned to the same agent, an agent can be assigned both bindings 2 and 3, provided that it has the capabilities to satisfy the corresponding parts of the formula. In addition, depending on the final assignments, the agents may need to synchronize with one another to perform parts of the task. For example, agents assigned with any subset of bindings $\{1,2,3\}$ need to synchronize their respective actions to satisfy $\varphi^\psi_1$.





\section{Control Synthesis for \ltlpsi}
\noindent \textbf{Problem statement:} Given $n$ heterogeneous agents $A = \{A_1,..., A_n\}$ and a task $\varphi^\psi$ in LTL$^{\psi}$, find a team of agents $\hat{A} \subseteq A$, their binding assignments $R_{\hat{A}}$, and synthesize behavior $\sigma_j$ for each agent such that $(\sigma(0), R_{\hat{A}}) \models \varphi^\psi$. This behavior includes synchronization constraints for agents to satisfy the necessary collaborative actions.
We assume that each agent is able to wait in any state (i.e. every state in the agent model has a self-transition).

\begin{figure}[t]
    \centering            \includegraphics[width=0.65\columnwidth]{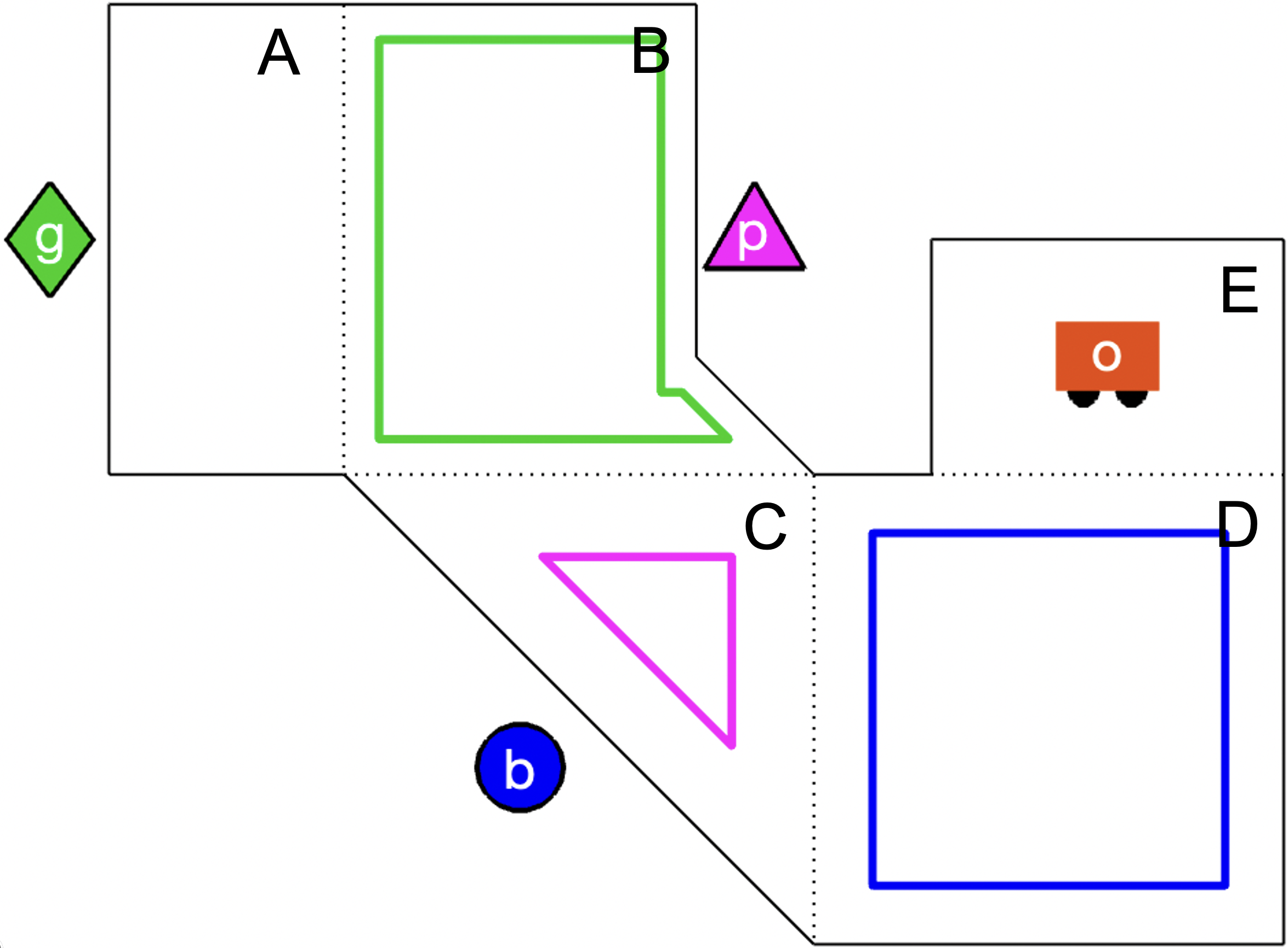}
\caption{ Agriculture environment and initial agent states. The green, blue, and pink agents are stationary; the orientation of their sensors are indicated by the colored boxes.
}
\label{fig:warehouse}
\end{figure}

\label{sec:example} 
\textbf{\textit{Example.}}
Consider a group of four agents $A = \{A_{green}$, $A_{blue}$, $A_{orange}$, $A_{pink}\}$  in a precision agriculture environment composed of 5 regions, as illustrated in Fig. \ref{fig:warehouse}. $A_{orange}$ is a mobile robot manipulator, such as Harvest Automation's HV-100, while the other agents are stationary with different onboard sensing capabilities.
The set of all capabilities is $\Lambda = \{$$\lambda_{\mathit{area\_j}}$,$\lambda_{motion}$, $\lambda_{arm}$, $\lambda_{UV}$, $\lambda_{moisture}$, $\lambda_{visual}$, $\lambda_{thermal}\}$, where $\forall j = \{green, blue, pink\}$, $\lambda_{\mathit{area\_j}}$ is agent $j$'s sensing area model. The green agent can orient its arm to reach either region A or B. The blue agent can orient its sensors to see one of three regions, B, C, or D; in order to reorient its sensors from regions B to D, its sensing range must first pass through region C. Similarly, the pink agent can orient its sensors to see either region A, B, or C, and its sensing range must pass through region B to get from regions A to C. The orange agent's ability to move between adjacent regions is represented by the capability $\lambda_{motion}$. Its sensing region is whichever region it is in. $AP_{arm} = \{\textit{pickup, dropoff, weed}\}$ is an abstraction of a robot manipulator that represents different actions the arm can perform, such as picking up soil samples or pulling weeds.  
$AP_{UV}$, $AP_{moisture}$, $AP_{visual}$, $AP_{thermal}$ all contain a single proposition representing a agent's ability to take UV measurements, soil moisture measurements, visual images, and thermal images, respectively. 
$\lambda_{arm}$ has more states (see Fig. \ref{fig:robot}b). Each agent may have distinct cost functions corresponding to individual capabilities.
 
The agent capabilities and label on the initial state are:\\
\noindent    $\Lambda_{green} = \{\lambda_{\mathit{area\_1}}, \lambda_{\mathit{arm}}\}, L(s_0) = \{region_B\}$ \\
\noindent    $\Lambda_{blue} = \{\lambda_{\mathit{area\_2}},\lambda_{\mathit{moisture}}, \lambda_{UV}\}, L(s_0) = \{region_D\}$ \\
\noindent    $\Lambda_{orange} = \{\lambda_{\mathit{motion}},\lambda_{\mathit{moisture}}, \lambda_{UV}, \lambda_{arm}\},L(s_0) = \{region_E\}$ \\
\noindent    $\Lambda_{pink} = \{\lambda_{\mathit{area\_4}},\lambda_{\mathit{thermal}}, \lambda_{\mathit{visual}}, \lambda_{\mathit{moisture}}, \lambda_{UV}\}$, $L(s_0) = \{region_C\}$

\noindent The team receives the task $\varphi^{\psi}$ (Eq.~\ref{eq:task}) and must determine a teaming assignment and behavior to satisfy the task. During execution, the agents must also synchronize with each other when necessary.

\section{Approach}

To find a teaming assignment and synthesize the corresponding synchronization and control, we first automatically generate a \buchi automaton $\mathcal{B}$ for the task $\varphi^\psi$ \ (Sec. \ref{sec:buchi_ltlpsi}). Each agent $A_j$ then constructs a product automaton $\mathcal{G}_j=A_j \times \mathcal{B}$ (Sec. \ref{sec:prod_aut}). For each binding $\rho \in AP_\psi$, it checks whether or not it can perform the task associated with that binding by finding a path to an accepting cycle in $\mathcal{G}_j$. Each agent creates a  copy of the \buchi automaton $\mathcal{B}_j$ pruned to remove any unreachable transitions and stores information about which combinations of binding assignments it can do. 

For parts of the task that require collaboration (e.g., when a transition calls for actions with bindings $\{1,2\}$ and $r_{green} = \{1,2\}, r_{blue} = \{2\}$), we need agents to synchronize. Thus, we synthesize behavior that allows for parallel execution while also guaranteeing that the team's overall behavior satisfies the global specification. 

To find a team of agents that can satisfy the task and their assignments, we need to guarantee that 1) every binding is assigned to at least one agent and 2) the agents synchronize for the collaborative portions of the task. 
To do so, we first run a depth-first search (DFS) to find a path through the $\mathcal{B}$ to an accepting cycle in which 
there exists a team of agents such that for every transition in the path, every proposition in $\APbinding$ is assigned to at least one agent (Sec. \ref{sec:team_dfs}). Each agent then synthesizes behavior to satisfy this path and communicates to other agents when synchronization is necessary.

\subsection{\buchi Automaton for an \ltlpsi\ Formula}\label{sec:buchi_ltlpsi}

When constructing a \buchi automaton for an \ltlpsi\ specification, we automatically rewrite the specification such that the binding propositions are only over individual atomic proposition $\pi \in AP_\varphi$ (i.e. the formula is composed of $\pi^\rho$).
For instance, the formula $(\neg pickup \ \mathcal{U} \ region_A)^{1 \vee 2}$ is rewritten as $(\neg pickup^1 \ \mathcal{U} \ region_A^1) \vee (\neg pickup^2 \ \mathcal{U} \ region_A^2).$ 

In our running example, we rewrite the formula in Eq. \ref{eq:task_part1} as
\begin{align}\label{eq:task_exp}
&\Diamond(region_B^2 \wedge moisture^2 \wedge UV^2 \\ &\wedge region_B^3 \wedge moisture^3 \wedge UV^3 \wedge region_A^1 \wedge pickup^{1}) \nonumber
\end{align}

\noindent\textbf{Remark 4.}
    In rewriting the specification, negation follows bindings in the order of operations. For example, $\neg pickup^{1\wedge 2} = \neg pickup^{1} \wedge \neg pickup^{2}$, and $\neg( pickup^{1\wedge 2}) = \neg (pickup^{1} \wedge pickup^{2}) = \neg (pickup^{1}) \vee \neg( pickup^{2})$. 

From $\APltl$ and $\APbinding$, we define the set of propositions $AP_{\varphi}^{\psi}$, where $\forall \pi \in \APltl$ and $\forall \rho \in \APbinding$, $\pi^\rho \in AP_{\varphi}^{\psi}$. Given $AP_{\varphi}^{\psi}$, we automatically translate the specification into a \buchi automaton using Spot \cite{spot}. 

To facilitate control synthesis, we transform any transitions in the \buchi automaton labeled with disjunctive formulas into disjunctive normal form (DNF).
We then replace the transition labeled with a DNF formula containing $\ell$ conjunctive clauses with $\ell$ transitions between the same states, each labeled with a different conjunction of the original label.

{In general, when creating a \buchi automaton from an LTL formula $\varphi$, $w\in\Sigma_{\mathcal{B}}$ are Boolean formulas over $AP_\varphi$, the atomic propositions that appear in $\varphi$, as seen in Fig. \ref{fig:buchi}. In the following, for creating the product automaton, we use an equivalent representation, where {$\Sigma_{\mathcal{B}} = 2^{AP_{\varphi}^{\psi}}\times 2^{AP_{\varphi}^{\psi}}$} and $w=(\sigma_T,\sigma_F)\in\Sigma_{\mathcal{B}}$ contains the set of propositions that must be true, $\sigma_T$, and the set of propositions that must be false, $\sigma_F$, for the Boolean formula over a transition to evaluate to True. These sets are unique in our case since each transition is labeled with a conjunctive clause (i.e. no disjunction).  Note that $\sigma_T\cap\sigma_F = \emptyset$ and $\sigma_T\cup\sigma_F\subseteq AP_\varphi$; propositions that do not appear in $w$ can have any truth value.}

Given a \buchi automaton for an \ltlpsi\ specification $\mathcal{B}$, we define the following functions:




\begin{figure*}[t]
    \centering
    \includegraphics[width=\textwidth]{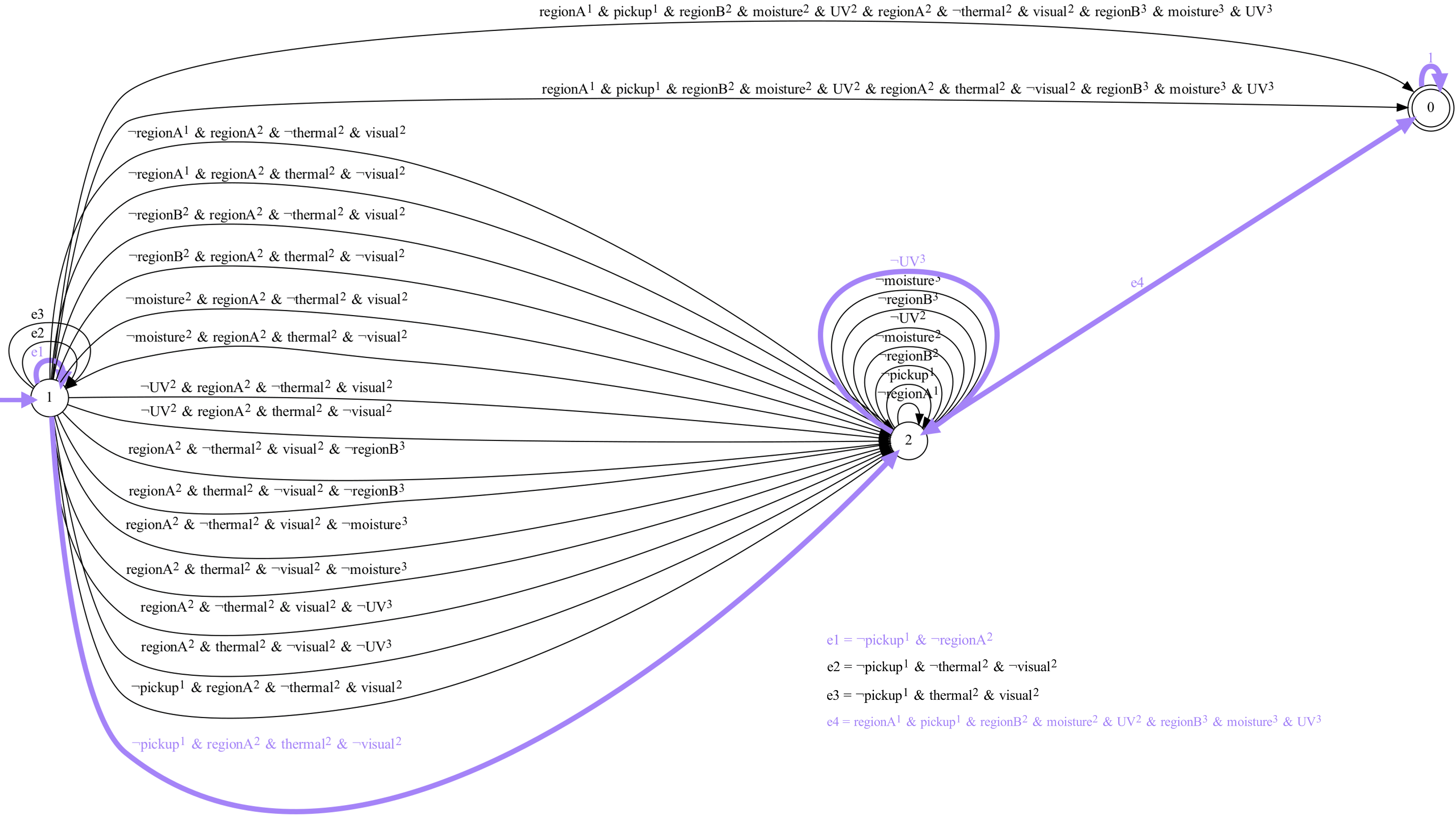}
    \caption{$\mathcal{B}$ for $\varphi^\psi$ (Eq. \ref{eq:task}). 
    The purple transitions illustrate a possible accepting trace.}
    \label{fig:buchi}
\end{figure*}

    

    
    
    
    
    

\noindent \textbf{Definition 1} (Binding Function).
$\mathfrak{B}: \Sigma_{\mathcal{B}} \rightarrow 2^{ AP_\psi}$ such that for $\sigma{=(\sigma_T,\sigma_F)} \in \Sigma_{\mathcal{B}}, \mathfrak{B}(\sigma) \subseteq \APbinding$ is the set $\{\rho \in\APbinding \ | \ \exists \pi^\rho \in {\sigma_T \cup \sigma_F\}}$.
Intuitively, it is the set of bindings that appear in label $\sigma$ of a \buchi transition.

\noindent \textbf{Definition 2} (Capability Function).
$\mathfrak{C}: \Sigma_{\mathcal{B}} \times \APbinding \rightarrow 2^{\APltl} \times 2^{\APltl}$ such that for $\sigma = (\sigma_T, \sigma_F) \in \Sigma_{\mathcal{B}}, \rho \in \APbinding$, $  \mathfrak{C}(\sigma, \rho) = (C_T, C_F)$, where $C_T = \{\pi \in\APltl \ | \ \exists \pi^\rho \in \sigma_T\}$ and $C_F = \{\pi \in\APltl \ | \ \exists \pi^\rho \in \sigma_F\}$. Here, $C_T$ and $C_F$ are the sets of action propositions that are True/False and appear with binding $\rho$ in label $\sigma$ of a \buchi transition.


\subsection{Agent Behavior for an \ltlpsi\ Specification}
\label{sec:prod_aut}

To synthesize behavior for an agent, we find an accepting trace in its product automaton $\mathcal{G}_j = A_j \times \mathcal{B}$, where $A_j = (S,s_0, AP_j, \gamma,L,W)$ is the agent model, and $\mathcal{B}= (Z,  z_0, \Sigma_{\mathcal{B}}, \delta_{\mathcal{B}}, F)$ is the \buchi automaton.

Since the set of propositions of $A_j$ may not be equivalent to the set of propositions of $\mathcal{B}$, we borrow from the definition of the product automaton in~\cite{Fang2022}. We first define the following function:

\noindent \textbf{Definition 3} (Binding Assignment Function).
Let $q = (s, z)$, $q' = (s',z')$, $\sigma=(\sigma_T, \sigma_F) \in \Sigma_{\mathcal{B}}$. Then $\mathfrak{R}(q, \sigma, q') = \{r \in 2^{\APbinding} \setminus \emptyset \ | \ \forall \rho \in r, (C_T, C_F)= \mathfrak{C}(\sigma, \rho),  \bigcup_{\rho \in r} C_T \subseteq L(s')$ and $\bigcup_{\rho \in r} C_F \cap L(s') = \emptyset \}$.

Intuitively, $\mathfrak{R}$ outputs all possible combinations of binding propositions that the agent can be assigned for a transition $(q,\sigma, q')$. An agent can be assigned $\rho$ if and only if the agent's next state $s'$ is labeled with all the action and motion propositions $\pi\in\APltl$ that appear in $\sigma_T$ as $\pi^\rho$, and all the propositions $\pi\in\APltl$ that appear in $\sigma_F$ as $\pi^\rho$ are not part of the state label (i.e. the agent is not performing that action). 
If a proposition $\pi^\rho$ is in $\sigma_F$ and $\pi$ is not in $AP_j$ (e.g. $scan^1 \in \sigma_F$ and the agent does not have $\lambda_{scan}$), the agent may be assigned $\rho$. Note that $r$ may include any binding propositions that are not in $\sigma$, since there are no actions required by those bindings in that transition. For example, if $\sigma = (\{scan^1\}, \{pickup^2\})$ and $AP_\psi = \{1,2,3\}$, then $\{3\}$ will be in the set $ \mathfrak{R}(q,\sigma, q')$ for all $q, q'$.  


Given $A_j$ and $\mathcal{B}$, we define the product automaton $\mathcal{G}_j = A_j \times \mathcal{B}$:

\noindent \textbf{Definition 4} (Product Automaton).
The product automaton $\mathcal{G}_j = (Q, q_0, AP_j,\delta_\mathcal{G},L_\mathcal{G}, W_\mathcal{G}, F_\mathcal{G})$, where 

\begin{itemize}[leftmargin=*]
    \item $Q = S \times Z$ is a finite set of states
    \item $q_0 = (s_0, z_0) \in Q$ is the initial state
    \item $\delta_\mathcal{G}\subseteq Q\times Q$ is the transition relation, where for $q = (s, z)$ and $q' = (s',z')$, $(q,q')\in \delta_\mathcal{G}$ if and only if $(s,s') \in \gamma$ and $\exists \sigma \in \Sigma_{\mathcal{B}}$ such that $(z, \sigma, z') \in \delta_{\mathcal{B}}$ and $\mathfrak{R}(q, \sigma, q') \neq \emptyset$ 
    


    \item $L_\mathcal{G}$ is the labeling function s.t. for $q = (s, z)$, $L_\mathcal{G}(q)\!=\!L(s)\!\subseteq\!AP_j$
    \item $W_\mathcal{G}: \delta_{\mathcal{G}} \rightarrow \mathbb{R}_{\ge 0} $ is the cost function s.t. for $(q, q') \in \delta_{\mathcal{G}}$, $q = (s, z)$, $q' = (s',z')$, $W_\mathcal{G}((q, q'))=W((s, s'))$
    \item $F_\mathcal{G} = S \times F$ is the set of accepting states

\end{itemize}

\begin{figure}[b]
    \centering            \includegraphics[width=0.9\columnwidth]{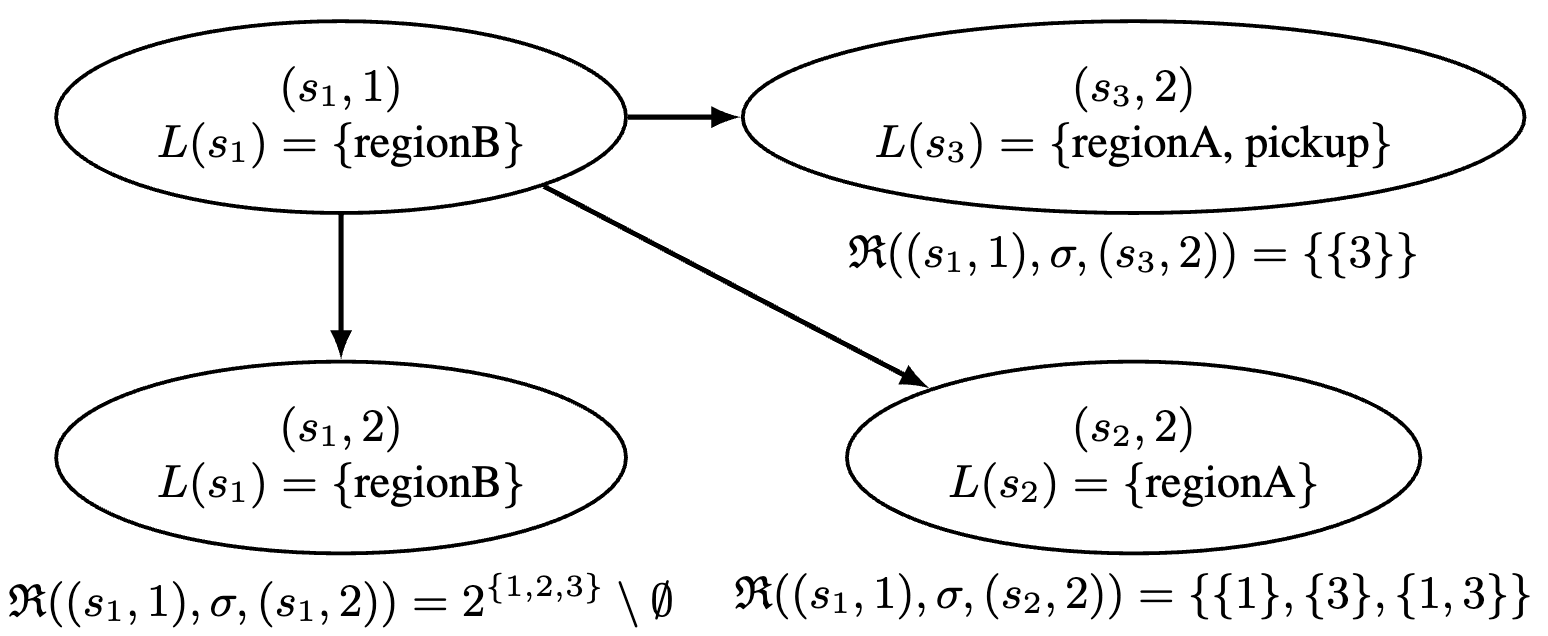}
\caption{A small portion of $\mathcal{G}_{green}$
}
\label{fig:prod_aut}
\end{figure}

\textbf{\textit{Example.}}
Fig. \ref{fig:prod_aut} depicts a small portion of $\mathcal{G}_{green}$; for the self-transition in $\mathcal{B}$ that is labeled with $\sigma=(\emptyset$, $\{pickup^1,$ $region_A^2\})$ (labeled as $e1$ in Fig. \ref{fig:buchi}), and for states in $A_{green}$  where $L(s_1) = \{region_B\}$, $L(s_2) = \{region_A\}$, $L(s_3) = \{region_A, pickup\}$, then the possible binding assignments are $\mathfrak{R}((s_1,1), \sigma, (s_1,2)) = 2^{\{1,2,3\}} \setminus \emptyset$ and  $\mathfrak{R}((s_1,1), \sigma, (s_2,2)) = \{\{1\},\{3\},\{1,3\}\}$. When the agent is in $s_3$, it cannot be assigned either bindings 1 or 2, but since no propositions appear with binding 3 in $\sigma$, $\mathfrak{R}((s_1,1), \sigma,(s_3,2)) = \{\{3\}\}$.


\subsection{Finding Possible Individual Agent Bindings} \label{sec: indiv_behavior}




To construct a team, we first reason about each agent and the sets of bindings it can perform. For example, for a formula $region_A^1 \wedge region_B^2$, an agent may be assigned $r_j=\{1\}$ or  $r_j=\{2\}$ but not $r_j=\{1,2\}$, since it cannot be in two regions at the same time. 

To find the set of possible binding assignments $R_j \subseteq 2^{\APbinding}$, we search for an accepting trace in $\mathcal{G}_j$ for every binding assignment $r_j\in  2^{\APbinding}$. We start from the full set of bindings $r_j=\APbinding$. Given an assignment $r_j$ to check, we find an accepting trace in $\mathcal{G}_j$ such that for all transitions $(q, q')$ in the trace, $r_j \subseteq \mathfrak{R}(q, \sigma, q')$. This ensures that the agent can satisfy its binding assignment for the entirety of its execution (i.e. $r_j$ does not change). Since every subset of a binding assignment $r_j$ is itself a possible binding assignment, if the agent can be assigned all $m = |\APbinding|$ bindings, then we know it can also be assigned every possible subset of $m$. If not, we check the ${m \choose m-1}$ combinations, and continue iterating until we have determined the agent's ability to perform every combination of the $m$ bindings.


Once an agent determines its possible binding assignments $R_j$, it creates the \buchi automaton $\mathcal{B}_j$ by removing any transition in $\mathcal{B}$ that cannot be traversed by any assignment in $R_j$. {In our example (Fig. \ref{fig:buchi}), each agent can be assigned at least one binding over every transition in $\mathcal{B}$. Thus, $\forall j \in \{green, blue, orange, pink\}, \mathcal{B}_j = \mathcal{B}$.}



\subsection{Agent Team Assignment} \label{sec:team_dfs}

A team of agents can perform the task if 1) all the bindings are assigned, with each agent maintaining the same binding assignment for the entirety of the task,
and 2) the agents satisfy synchronization requirements. For a viable team, the agents' control follows the same path in the \buchi automaton $\mathcal{B}$ to an accepting cycle. We perform DFS over $\mathcal{B}$ to find an accepting trace (Alg. \ref{algo:buchi_dfs}), where  each tuple in $stack$ contains the current edge $(z, \sigma, z')$, the current team of agents $R_{\hat{A}}$, and the path traversed so far $\beta_{\hat{A}}$.

We initialize the team with all agents $A_j$ and all possible binding assignments $R_j$, and each path $\beta_{\hat{A}}$ starts from state $z_0$ of $\mathcal{B}$. When checking a transition $(z, \sigma, z')$, we remove any agent $j$ if $ \forall ((s,z),(s',z')) \in \delta_{\mathcal{G}_j}$, there are no possible binding assignments it can satisfy. This is done by checking each agent's pruned \buchi automaton $\mathcal{B}_j$ in $\textsc{update\_team}$  (line \ref{line:update_team}). We want the agent's behavior to satisfy not only the current transition, but also the entire path with a consistent binding assignment. Thus, we update possible bindings ($\textsc{update\_bindings}$, lines \ref{line:update_combos1}-\ref{line:update_combos2}). 

To guarantee the overall team behavior, we need to ensure agents are able to ``wait in a state" before they synchronize, as they may reach states at different times. This means that each state in the trace must have a corresponding self-transition. Thus, for every $(z, \sigma, z')$ that we add to the path in which $z \neq z'$, the next edge to traverse must be a self-transition from $z'$ to itself; the same holds vice-versa. In line \ref{line:check_self}, we check if the current transition is self-looping or not, and add subsequent transitions into the stack accordingly. If there is no self-transition on $z'$ (i.e. $(z', \sigma, z') \notin \delta_{\mathcal{B}}$), then we do not consider $z'$ to be valid and do not add it to the path.



Once we find a valid path to an accepting cycle, we parse it into
$\beta$, the path without self-transitions, and $\delta_{self}$, which contains the corresponding self-transition for each state in the path. Fig. \ref{fig:buchi} shows a valid path in $\mathcal{B}$ for the example in Sec. \ref{sec:example} and the corresponding team assignment $\hat{A}=\{A_{green}, A_{blue}, A_{orange}, A_{pink}\}$ and bindings $r_{green} = \{1\}$, $r_{blue} = \{3\}, r_{orange} = \{1\}, r_{pink} = \{2,3\}$. 
Note that we find a valid path rather than a globally optimal one. However, the algorithm is complete; it will find a feasible path if one exists. 

\setlength{\textfloatsep}{0pt}
\begin{algorithm}[t]
    \SetKwInOut{Input}{Input}
    \SetKwInOut{Output}{Output}
    \SetKwProg{Initialization}{Initialization}{}{}
    \Input{$A=\{A_1, A_2, ..., A_n\}$, $R = \{R_1, R_2, ..., R_n\}$, $\mathcal{B}$, $\{ \mathcal{B}_1, \mathcal{B}_2 ..., \mathcal{B}_n \}$}
    \Output{$\beta$, $\delta_{self}$, $\hat{A} \subseteq A$, $R_{\hat{A}}$}
    
    $stack = \emptyset$, $visited = \emptyset$  \\
    \For{$e \in \{(z, \sigma, z') \in \delta_{\mathcal{B}} \ | \ z = z_0 \}$ }{
        $stack = stack \cup \{(e, R, [e])\}$ 
    }

    \While{$stack \neq \emptyset$}{
        $((z, \sigma, z'), R_{\hat{A}}, \beta_{\hat{A}}) = stack.pop()$ \\
        
        \If{$(z, \sigma, z') \not\in visited$}{
        $visited = visited \cup (z, \sigma, z')$ \\

        $R_{\hat{A}} = \textsc{update\_team}((z, \sigma, z'), \{ \mathcal{B}_1, ..., \mathcal{B}_n \})$ \label{line:update_team}
        
        
        \For{$R_j \in R_{\hat{A}}$}{ \label{line:update_combos1}
            $R_j' = \textsc{update\_bindings}(R_j, (z, \sigma, z'))$  \\
            \If{$R_j' = \emptyset$}{
                $R_{\hat{A}} = R_{\hat{A}} \setminus R_j$
            }
            \Else{$R_{\hat{A}} = (R_{\hat{A}} \setminus R_j) \cup R_j'$}

        }
        \label{line:update_combos2}

        \If{$\bigcup_j (R_j \in R_{\hat{A}}) = \APbinding$}{
            
            \If{$z' \in F$}{ 
            $\beta, \delta_{self} = \textsc{parse\_path}(\beta_{\hat{A}})$ \\
            \Return $\beta, \delta_{self}, R_{\hat{A}}$}
            
            
            $E = \{(z',\sigma', z'') \in \delta_{\mathcal{B}}\}$ \label{line:E_z}
            
            \For{$(z',\sigma', z'') \in  E$
                }{
                \If{$(z = z'  \text{ and } z'\neq z'')$ or 
                 $(z \neq z'  \text{ and } z'= z'')$ \label{line:check_self}}{
                    $stack = stack \cup \{ \left( (z',\sigma', z''), R_{\hat{A}}, [\beta_{\hat{A}} \ (z',\sigma', z'')]\right)\}$
                    }
                }
            
            }

            
        
        
        }
    }
\caption{Find Accepting Trace for Agent Team}
\label{algo:buchi_dfs}
\end{algorithm}

\subsection{Synthesis and Execution of Control and Synchronization Policies} \label{sec: sync_behavior}

Given an accepting trace $\beta$ through $\mathcal{B}$ and the corresponding self-transitions $\delta_{self}$ that are valid for all agents in $R_{\hat{A}}$, we synthesize control and synchronization for each agent such that the overall team execution satisfies $\beta$ (Alg. \ref{algo:synthesis}). 
For each transition $(z, \sigma, z')$ in $\beta$, we find $\overline{R}$, which contains the binding assignments of all agents that require synchronization at state $z'$.
Agent $j$ participates in the synchronization step if $r_j$ contains a binding $\rho$ that is required by $\sigma$ and is not the only agent assigned bindings from $\sigma$ (line \ref{line:Rbar}).

Subsequently, agent $j$ finds an accepting trace in $\mathcal{G}_j$ that reaches $z'$ with minimum cost, following self-transitions stored in $\delta_{self}$ if necessary. As it executes this behavior, it communicates with other agents the tuple $p$, which contains 1) its ID, 2) the state $z'$ it is currently going to, and 3) if it is ready for synchronization (line \ref{line:execute}). If no synchronization is required (line \ref{line:nosync}), the agent can simply execute the behavior. Otherwise, to guarantee that the behavior does not violate the requirements of the task, the agent executes the synthesized behavior up until the penultimate state, $z_{wait}$. 

When the agent reaches $z_{wait}$, it signals to other agents that it is ready for synchronization. Since all agents know the overall teaming assignment, the agent continues to wait in state $z_{wait}$ until it receives a signal that all other agents in $\overline{R}$ are ready (line \ref{line:check_sync}). These agents then move to the next state in the behavior simultaneously. Agent $j$ continues synthesizing behavior through $\beta$ until synchronization is necessary again, and this process is repeated. 
\begin{algorithm}[t]
    \SetKwInOut{Input}{Input}
    \SetKwInOut{Output}{Output}
    \SetKwProg{Initialization}{Initialization}{}{}
    \Input{$\mathcal{G}_j$, $r_j$, $R_{\hat{A}}$, $\beta$, $\delta_{self}$}
    \For{$(z,\sigma, z') \in \beta$}{
    $b_j = \textsc{find\_behavior}(\mathcal{G}_j, r_j, (z,\sigma, z'), \delta_{self})$ \\
    $\overline{R} = \{r_k \in R_{\hat{A}} \ | \ r_k \cap \mathfrak{B}(\sigma) \neq \emptyset \}$ \label{line:Rbar}
    \If{$r_j \not \in \overline{R}$ or $\overline{R}=\{r_j\}$ \label{line:nosync}}{
    $p = ()$ \\
    $\textsc{execute}(b_j, p)$
    }
    \Else{
    $p = (j, z', 0)$, $\ell = length(b_j)$ \\
    $\textsc{execute}(b_j[1:\ell-1], p)$ 
    \label{line:execute}\\
    $z_{wait} = b_j[\ell-1]$, $P = \{j\}$ \label{line:wait} \\
    \While{$\bigcup_{i \in P} (r_i \in \overline{R}) \neq \mathfrak{B}(\sigma)$}{
       $p = (j, z', 1)$ \\
        $\textsc{execute}(z_{wait}, p)$ \\
        $P = j \cup \{ k \ | \ (k, z', 1) \in \textsc{receive}()\}$ \label{line:check_sync}   
    }
    $\textsc{execute}(b_j[\ell])$
    }
    }     
\caption{Synthesize an Agent's Behavior}
\label{algo:synthesis}
\end{algorithm}

 \section{Results and Discussion}

\begin{figure}[h]
    \centering
    \includegraphics[width=0.9\columnwidth]{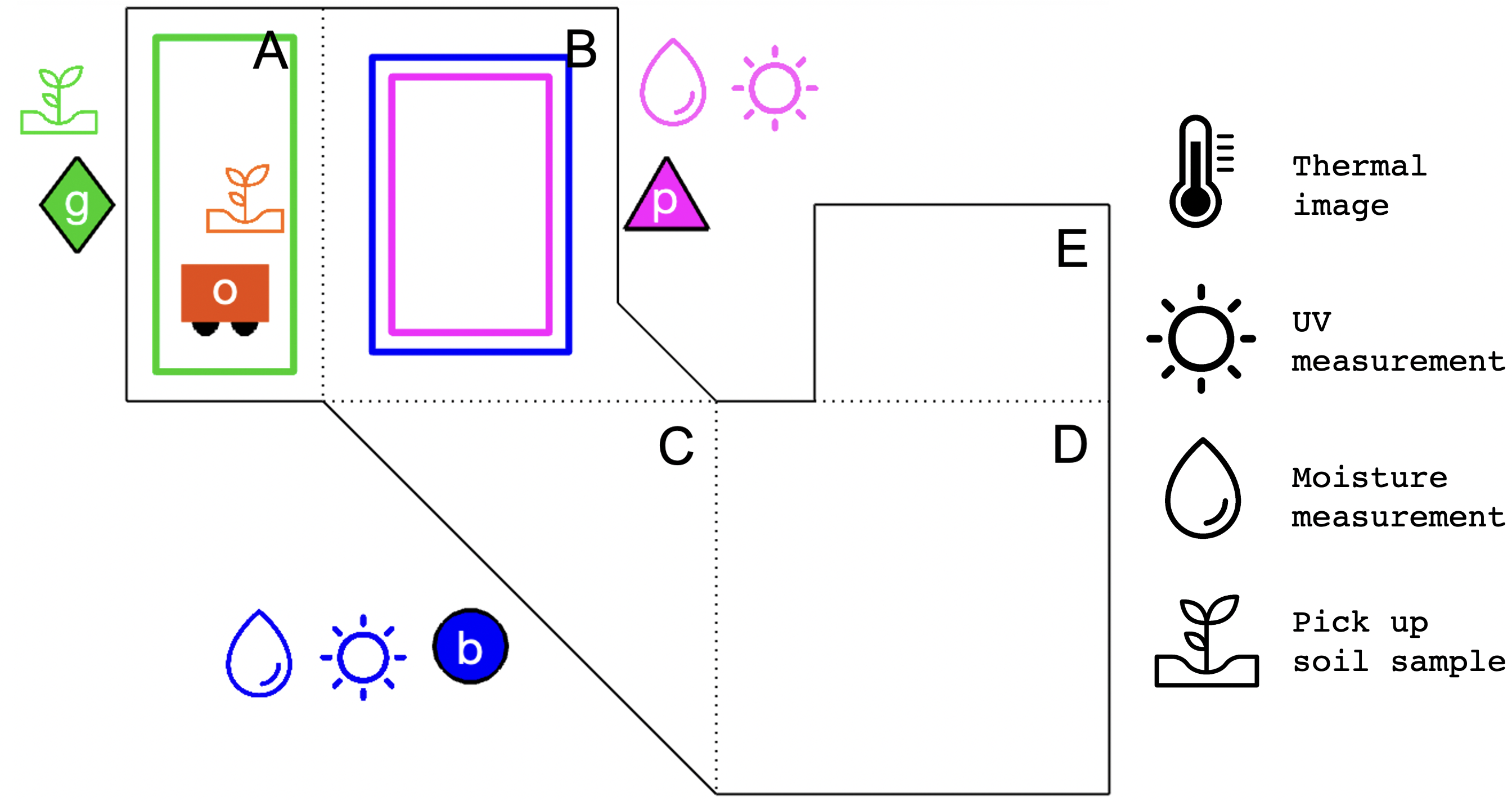}
    \caption{The final step in the synchronized behavior of the agent team with their corresponding actions.}
    \label{fig:results_behavior}
\end{figure}

Fig. \ref{fig:results_behavior} shows the final step of the synchronized behavior of the agent team for the example in Section \ref{sec:example}, where $\hat{A}=\{$$A_{green}$, $A_{blue}$, $A_{orange}$, $A_{pink}\}$ with binding assignments $r_{green} = \{1\}$, $r_{blue} = \{3\}, r_{orange} = \{1\}, r_{pink} = \{2,3\}$. A simulation of the full behavior is shown in the accompanying video. 

 
 \noindent \textbf{Optimizing teams:} Our synthesis algorithm can be seen as a greatest fixpoint computation, where we start with the full set of agents and remove those that cannot contribute to the task. 
 As a result, the team may have redundancies, i.e. agents can be removed while still ensuring the overall task will be completed; this may be beneficial for robustness. Furthermore, we can choose a sub-team to optimize different metrics, as long as the agent bindings assignments still cover all the required bindings. 
For example, minimizing the number of bindings per agent could result in $\hat{A} = \{A_{green}, A_{blue}, A_{pink}\}$, $r_{green} = \{1\},r_{blue} = \{3\}, r_{pink} = \{2\}$; minimizing the number of agents results in $\hat{A} = \{A_{green}, A_{pink}\}$, $r_{green} = \{1\}, r_{pink} = \{2,3\}$. 
 
To illustrate other possible metrics, we consider a set of 20 agents and create a team for the specification in Eq.~\ref{eq:task}. Their final binding assignments and costs are shown in Table \ref{table:20agents}. Minimizing cost results in a team $\hat{A} = \{A_7, A_{11}\}$. Minimizing cost while requiring each binding to be assigned to two agents results in $\hat{A} = \{A_4, A_7, A_{11}, A_{16}\}$. 


\begin{table*}[t]
\centering 
    \begin{tabular}{ |ccc|ccc|ccc|ccc|ccc| } 
     \hline
     Agent & $r_j$ & $cost$ & Agent & $r_j$ & $cost$ & Agent & $r_j$ & $cost$ & Agent & $r_j$ & $cost$ & Agent & $r_j$ & $cost$ \\
     \hline
     {1} & {1} & 1.2 & 
     {5} & {3} & 2.75 & 
     {9} & {3} & 2.6 & 
     {13} & {3} & 2.0 &  
     {17} & {2,3} & 3.275 \\
     \hline
     {2} & {3} & 1.0 & 
     {6} & {1} & 0.95 & 
     {10} & {1} & 2.8 & 
     {14} & {1} & 1.2 &
     {18} & {3} & 2.55 \\ 
     \hline
     {3} & {1} & 1.2 & 
     {7} & {1} & 0.65 & 
     {11} & {2,3} & 0.9 & 
     {15} & {3} & 1.1 &
     {19} & {1} & 1.9 \\ 
     \hline
     {4} & {2,3} & 1.3 & 
     {8} & {1} & 1.0 & 
     {12} & {2,3} & 1.825 & 
     {16} & {1} & 0.775 &
     {20} & {2,3} & 2.35 \\ 
     \hline
    \end{tabular} 
    \caption{Example teaming assignment with 20 robots }
    \label{table:20agents}
\end{table*}

\noindent \textbf{Computational complexity:} 
{The control synthesis algorithm (Alg.~\ref{algo:synthesis}) is agnostic to the number of agents, since each agent determines its own possible bindings assignments and behavior.} For the team assignment (Alg. \ref{algo:buchi_dfs}), since it is a DFS algorithm, we need to store the agent team and their possible binding assignments as we build an accepting trace. Thus, it has both a space and time complexity of $O(|E|*2^{m}*n)$, where $|E|$ is the number of edges in $\mathcal{B}$, $m$ is the number of bindings, and $n$ is the number of agents.

\begin{figure}
     \centering
     \begin{subfigure}[t]{\columnwidth}
         \centering
         \includegraphics[width=0.74\textwidth]{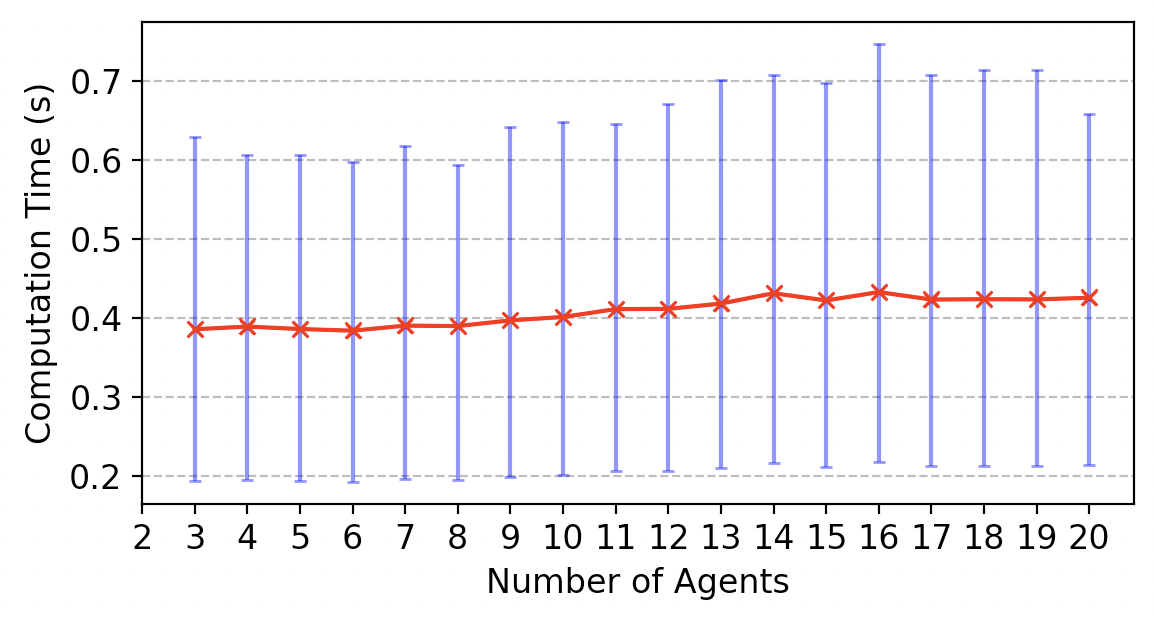}
         \caption{}
         \label{fig:vary_robots}
     \end{subfigure}
    \centering
     \begin{subfigure}[t]{\columnwidth} 
         \centering
         \includegraphics[width=0.74\textwidth]{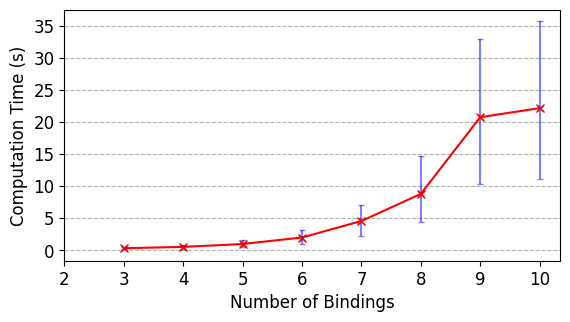}
         \caption{}
         \label{fig:vary_bindings}
     \end{subfigure}
    \caption{Computation time when increasing the number of agents (\subref{fig:vary_robots}) and the number of bindings (\subref{fig:vary_bindings}). The error bars represent min/max values.}
    \label{fig:sim_results}
\end{figure}

Fig. \ref{fig:vary_robots} shows the computation time of the synthesis framework (Sec. \ref{sec:prod_aut} -- \ref{sec:team_dfs}) for simulated agent teams in which we vary the number of agents from 3 to 20, running 30 simulations for each set of agents and randomizing their capabilities. The task for each
simulation is the example in Eq. \ref{eq:task}. We also ran simulations in which we increase the number of bindings from 3 to 10 and randomized the capabilities of 4 agents (Fig. \ref{fig:vary_bindings}). The variance in computation time is a result of the randomized agent capabilities, which affects the computation time of possible binding assignments (Sec. \ref{sec: indiv_behavior}). All simulations ran on a 2.5 GHz quad-core Intel Core i7 CPU.

  \noindent \textbf{Task expressivity with respect to other approaches:} 
 We compare \ltlpsi \ to other approaches that encode collaborative heterogeneous multi-agent tasks using temporal logic. 
 
\textit{Standard LTL:} One approach is to use LTL to express the task by enumerating all possible assignments in the specification. In our example, Eq. \ref{eq:task_part1} would be rewritten as:
\begin{align*}\label{eq:task_naive}
\varphi^{\psi}_1 = &(\Diamond((region_B^{green} \wedge moisture^{green} \wedge UV^{green}) \nonumber \\ 
\wedge  &(region_A^{blue} \wedge pickup^{blue}))) \nonumber \\
\vee &(\Diamond((region_B^{green} \wedge moisture^{green} \wedge UV^{green})  \nonumber \\
\wedge  &(region_A^{orange} \wedge pickup^{orange}))) 
\vee ...
\end{align*} 

\noindent where each agent has its own unique set of $AP$, denoted here by each proposition's superscript. As a result, the number of propositions increases exponentially with the number of agents. The task complexity also increases, as the specification must include all possible agent assignments. Another drawback of using LTL for such tasks is that the specification is not generalizable to any number of agents; it must be rewritten when the set of agents change.

\textit{LTL$^{\chi}$:} In~\cite{Luo2022}, tasks are written in LTL$^{\chi}$, where proposition $\pi_{i,j}^{k, \chi}$ is true if at least $i$ agents of type $j$ are in region $k$ with binding $\chi$. We can express $\varphi^\psi_1$ (Eq. \ref{eq:task_part1}) of our example as $\Diamond(\pi_{1,mois}^{regionB,2} \wedge \pi_{1,UV}^{regionB,2} \wedge \pi_{1,mois}^{regionB,3} \wedge \pi_{1,UV}^{regionB,3} \wedge  \pi_{1,arm}^{regionA,1})$. The truth value of $\pi_{i,j}^{k, \chi}$ is not dependent on any particular action an agent might take. LTL$^{\chi}$ can be extended to action propositions, but since an agent can only be categorized as one type, each type of agent must have non-overlapping capabilities (here, we have written the LTL$^{\chi}$ formula such that each type of agent only has one capability). In addition, $\varphi^\psi_2$ (Eq. \ref{eq:task_part2}) cannot be written in LTL$^{\chi}$ because the negation defined in our grammar cannot be expressed in  LTL$^{\chi}$. On the other hand, the negative  proposition $\neg \pi_{i,j}^{k, \chi}$ from~\cite{Luo2022} is equivalent to ``less than $i$ agents of type $j$ are in region $k$", which our logic cannot encode. 

\textit{Capability Temporal Logic (CaTL):} Tasks in CaTL \cite{Leahy2022} are constructed over tasks $T = (d, \pi, cp_T)$, where $d$ is a duration of time, $\pi$ is a region in AP, $(c_i, m_i) \in cp_T$ denotes that at least $m_i$ agents with capability $c_i$ are required. Similar to our grammar, CaTL allows agents to have multiple capabilities, but each task must specify the number of agents required. Since it is an extension of Signal Temporal Logic, tasks provide timing requirements, which our logic cannot encode. However, it does not include the concept of binding assignments; in our example $\varphi^\psi_1$ (Eq. \ref{eq:task_part1}), CaTL cannot express that we require the same agent that took a UV measurement to also take a thermal image. 
Ignoring binding assignments and adding timing constraints, $\varphi^\psi_1$ (Eq. \ref{eq:task_part1}) can be rewritten in CaTL as $\Diamond_{[0,10)}$$(T(0.1, region_B, $ $\{(moisture, 2), (UV, 2)\}) \wedge$  $T(0.5, $ $region_A,$ $\{(arm, 1)\})$. Each capability in CaTL is represented as a sensor and therefore cannot include more complex capabilities, such as a robot arm that can perform several different actions.
In addition, because CaTL requires the formula to be in positive normal form (i.e. no negation), we cannot express $\varphi^\psi_2$ (Eq. \ref{eq:task_part2}) in this grammar.

 \section{Conclusion}
We define a new task grammar for heterogeneous teams of agents and develop a framework to automatically assign the task to a (sub)team of agents and synthesize correct-by-construction control policies to satisfy the task. We include synchronization constraints to guarantee that the agents perform the necessary collaborations. 

In future work, we plan to demonstrate the approach on physical systems where we need to ensure that the continuous execution satisfies all the collaboration and safety constraints. In addition, we will explore different notions of optimality when finding a teaming plan, as well as increase the expressivity of the grammar by allowing reactive tasks where agents modify their behavior at runtime in response to environment events. 



\begin{acks}
This work is supported by the National Defense Science \& Engineering Graduate Fellowship 
(NDSEG) Fellowship Program.
\end{acks}



\bibliographystyle{ACM-Reference-Format} 
\bibliography{references}


\end{document}